\documentclass[]{llncs}
\usepackage{amsmath,amsfonts,svg}
\usepackage[T1]{fontenc}
\usepackage{graphicx}
\begin{document}
\title{Effect of Kernel Size on CNN-Vision-Transformer-Based Gaze Prediction Using Electroencephalography Data}
\titlerunning{ViT-CNN-Based Gaze Prediction Using EEG}
%
\author{Chuhui Qiu\inst{1}\orcidID{0009-0008-7780-1291} \and
Bugao Liang\inst{1}\orcidID{0009-0008-7705-1590} \and
Matthew L Key\inst{1}\orcidID{0009-0004-3984-8209}}
\authorrunning{C. Qiu et al.}
%
\institute{The George Washington University, Washington DC 20052, USA\\
\email{\{chqiu,bliang271,matthewlkey\}@gwmail.gwu.edu}}
\maketitle              
\begin{abstract}
In this paper, we present an algorithm of gaze prediction from Electroencephalography (EEG) data. EEG-based gaze prediction is a new research topic that can serve as an alternative to traditional video-based eye-tracking. Compared to the existing state-of-the-art (SOTA) method, we improved the root mean-squared-error of EEG-based gaze prediction to 53.06 millimeters, while reducing the training time to less than 33\% of its original duration. Our source code can be found at https://github.com/AmCh-Q/CSCI6907Project.

\end{abstract}
\section{Introduction}

Electroencephalography (EEG) is a non-invasive technique used to record the electrical activity generated by the brain. 
Owing to its relative accessibility, non-invasiveness, superior temporal resolution compared to other neuroimaging techniques such as positron emission tomography (PET) or functional magnetic resonance imaging (fMRI), EEG's potential extends to many different fields. One such application is the complimentary application in eye-tracking. As existing video-based eye-tracking methods rely on setting up fixed cameras and pointing them directly toward the subject's eyes, EEG-based eye-tracking may lead to a promising alternative solution that does not necessarily require fixed cameras within the subject's field-of-view.

EEGViT \cite{yang2023vit2eeg} is the current state-of-the-art (SOTA) model on EEG-based gaze prediction accuracy on the EEGEyeNet dataset \cite{kastrati2021eegeyenet}. It employs a hybrid transformer model fine-tuned with EEG data \cite{khan2022transformers,vaswani2017attention}.

\subsection{Research Question}
In this paper, we propose a method that answer the following questions:
\begin{itemize}
    \item In CNN-transformer hybrid models, how do different convolution kernel sizes over the EEG spatial features (channels) affect the accuracy of the CNN-transformer hybrid models?
    \item How does this compare against a convolution over all EEG channels?
\end{itemize}

\noindent By answering this question, we investigate the effects of convolution kernels on the CNN-transformer hybrid networks.

\section{Related Work}
Over the past decade, a variety of machine learning and deep learning algorithms have been utilized to process EEG data, resulting in significant advancements and applications in areas such as emotion recognition, motor imagery, mental workload evaluation, seizure detection, Alzheimer's disease classification, and sleep stage analysis, among others \cite{craik2019deep,roy2019deep,altaheri2023deep,gao2021complex,hossain2023status,key2024advancing,li2024enhancing,koome2023trends,murungi2023empowering,qu2022eeg4home,yi2022attention,dou2022time,zhou2022brainactivity1,wang2022eeg,qu2022time,qu2020identifying,qu2020using,qu2020multi,qu2018eeg,qu2019personalized}.
While EEG and Eye-tracking have each been studied individually for over a century, their combined use has only seen an increased interest in recent years with the aid of convolutional neural networks (CNNs) and transformers.

\subsection{Dataset}
\begin{figure}
    \centering
    \includegraphics[width=0.7\linewidth]{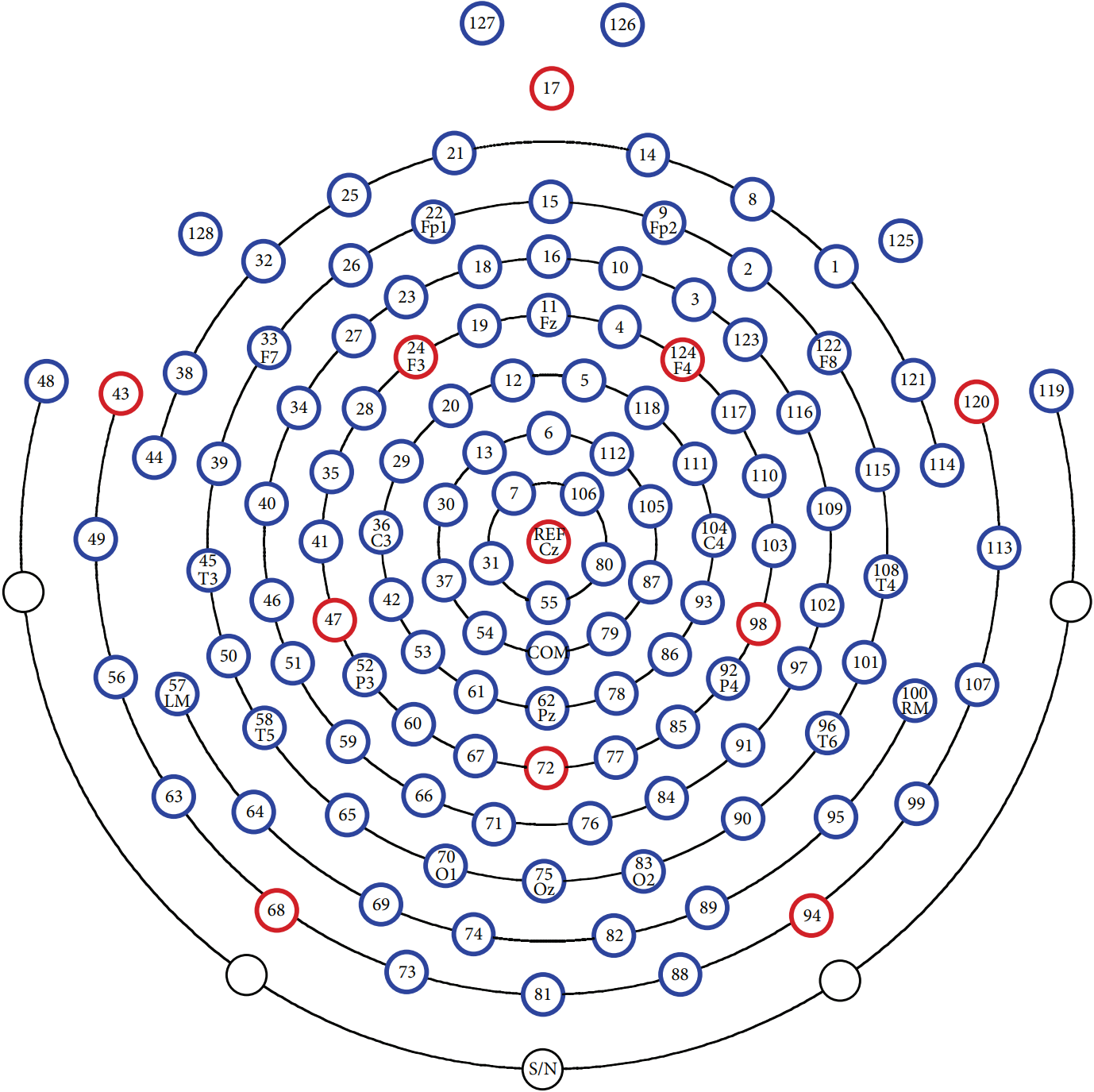}
    \caption{Electrode Layout of the 128-channel EEG Geodesic Hydrocel system \cite{bamatraf2016system}}
    \label{fig:hydrocel}
\end{figure}
\begin{figure}
    \centering
    \includegraphics[width=0.7\linewidth]{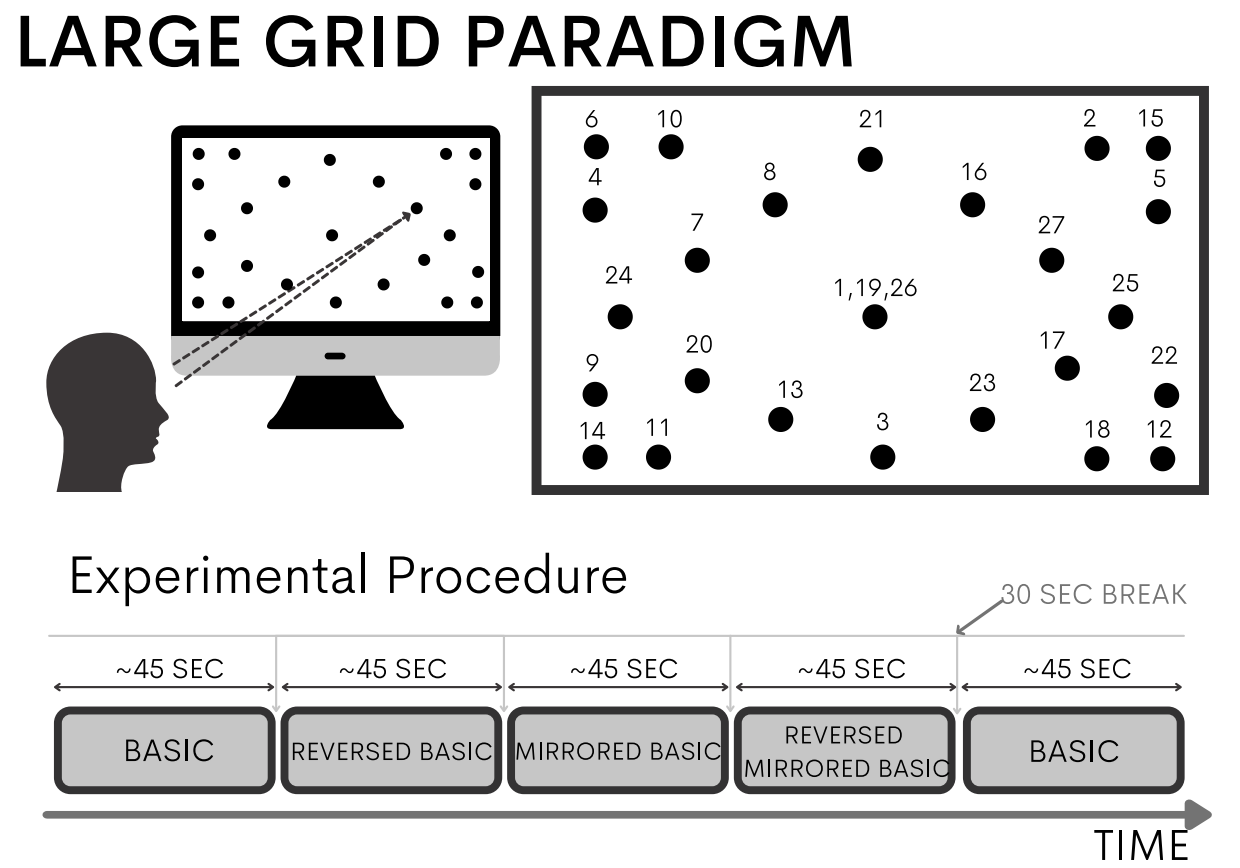}
    \caption{The Large Grid Paradigm of EEGEyeNet \cite{kastrati2021eegeyenet}}
    \label{fig:large_grid_paradigm}
\end{figure}
\begin{figure}
    \centering
    \includegraphics[width=0.7\linewidth]{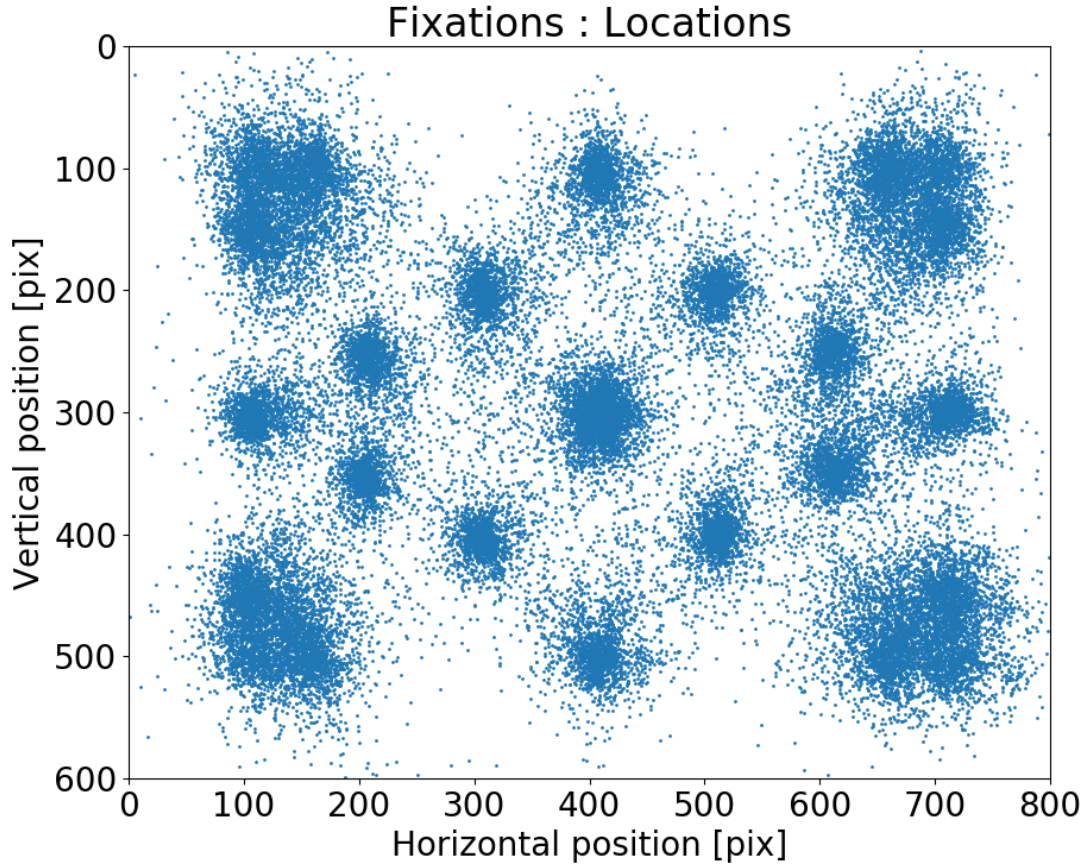}
    \caption{Distribution of the Fixation Positions in the Large Grid Paradigm \cite{kastrati2021eegeyenet}}
    \label{fig:fixations}
\end{figure}
The EEGEyeNet dataset \cite{kastrati2021eegeyenet} offers EEG and eye tracking data that were collected simultaneously as well as benchmarks for eye movement and gaze position prediction. The EEG data of EEGEyeNet are collected from 356 participants using a 128-channel EEG Geodesic Hydrocel system, where the EEG channels are individually numbered from 1 to 128 as shown in Figure \ref{fig:hydrocel}. An additional reference electrode in the center make up a total of 129 EEG channels in the raw dataset.
\subsubsection{Experimental Paradigm}
In one of EEGEyeNet's experimental paradigms, the participants are asked to fixate on specific dots on an "large grid" on the screen for a period as seen in Figure \ref{fig:large_grid_paradigm}. At the same time of recording EEG data, the participants' gaze positions are recorded. The gaze position distributions of 21464 samples can be seen in Figure \ref{fig:fixations}\cite{kastrati2021eegeyenet}.
\subsection{State-of-the-art}
\begin{figure}
    \centering
    \includegraphics[width=1\linewidth]{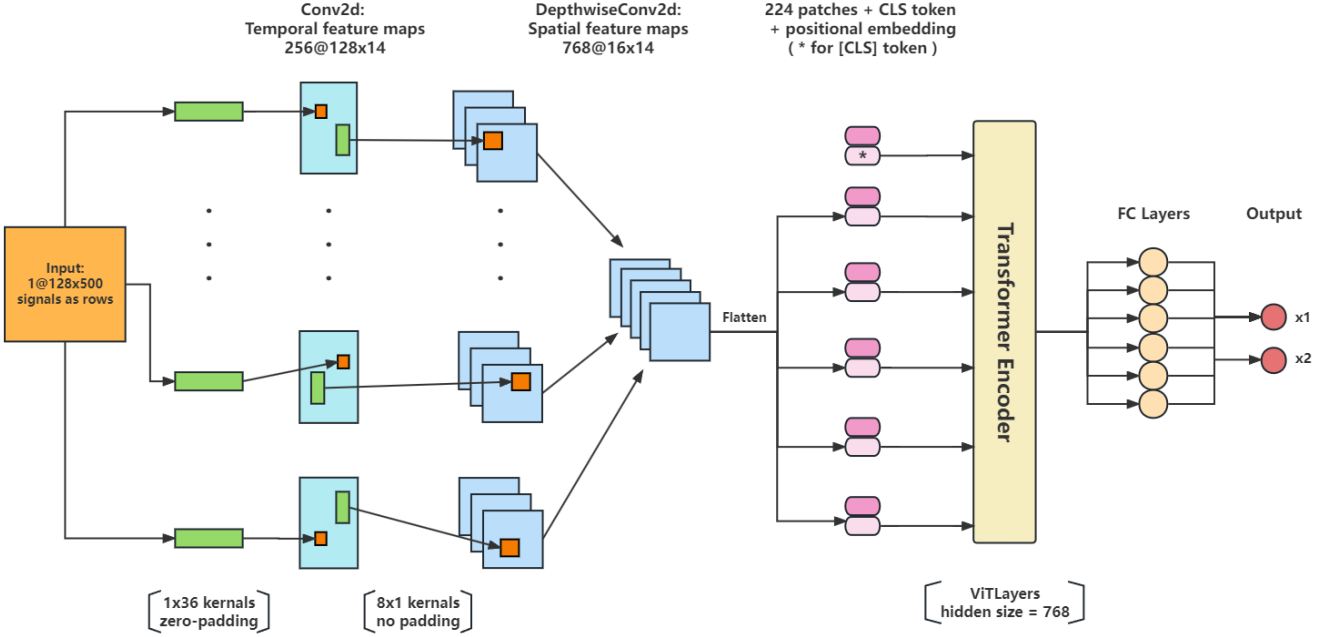
}
    \caption{EEGViT Model Architecture \cite{yang2023vit2eeg}}
    \label{fig:vit2eeg_arch}
\end{figure}
Since the publication of EEGEyeNet, several follow-up works have been made, often focusing on classification tasks (left-right or events such as blinking) \cite{wolf2022deep,xiang2022too,xiang2022vector}. The current state-of-the-art model in predicting gaze position is EEGViT \cite{yang2023vit2eeg}, a hybrid vision transformer model fine-tuned with EEG data as shown in Figure \ref{fig:vit2eeg_arch}. EEGViT combines a two-level convolution feature extraction method, previously proposed in EEGNet \cite{lawhern2018eegnet} and Filter Bank Common Spatial Patterns \cite{schirrmeister2017deep} which enables efficient extraction of spatial (EEG electrodes) features for each temporal (frequency) channel, and a vision transformer using the ViT-Base model \cite{dosovitskiy2020image} pre-trained with ImageNet \cite{deng2009imagenet,ridnik2021imagenet}, to achieve a reported RMSE of $55.4\pm 0.2$ millimeters on the EEGEyeNet dataset \cite{yang2023vit2eeg}.

\section{Experiment}
\subsection{Model}

\begin{table}
    \centering
    \caption{Detailed Description of Our Model Architecture}
    \label{tab:method_1_layers}
    \begin{tabular}{c | c}
        Layer & Description\\
        \hline
        0 & Input Size $129\times500\times 1$, Zero-padded to $129\times512\times 1$ on both sides\\
        1 & 256 Temporal Convolution size $1\times 16$ for Kernel and Stride, Batchnorm\\
        2 & 768 Spatial Convolution size $129\times 1$\\
        3 & ViT Model transformer, image size $129\times 32$, patch size $129\times 1$\\
        4 & Linear layer with 768 neurons on top of the final hidden CLS token\\
        5 & Linear layer with 1000 neurons, Dropout $p=0.1$\\
        6 & Linear layer with 2 neurons (output)
    \end{tabular}
\end{table}

\begin{figure*}
    \centering
    \includegraphics[
width=1\linewidth]{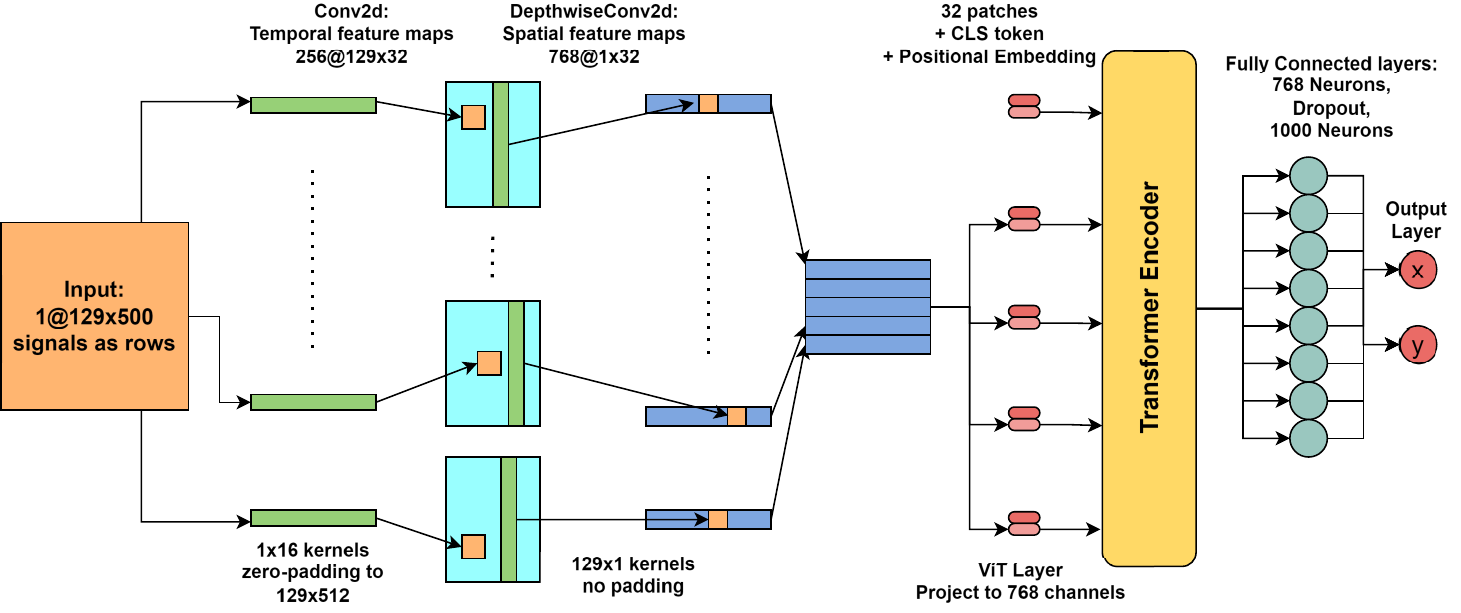}
    \caption{Our Model Architecture, modified from \cite{yang2023vit2eeg}}
    \label{fig:method_1_arch}
\end{figure*}

The architecture of our Method can be seen in Figure \ref{fig:method_1_arch} and Table \ref{tab:method_1_layers}. Similar to prior works \cite{lawhern2018eegnet,schirrmeister2017deep,yang2023vit2eeg}, we employ two convolution layers which filter the temporal and spatial (EEG channels) dimensions respectively.

In the first layer, a $1\times16$ kernel scans across the 1-second $129\times 500$ input which is zero-padded to $129\times 512$. The kernels effectively function as band-pass filters on the raw input signals. Our choice of $1\times 16$ kernel is smaller than that of EEGViT at $1\times 36$ \cite{yang2023vit2eeg} and that of EEGNet at $1\times 64$ \cite{lawhern2018eegnet}. This provides a greater resolution of temporal features to be learned. Batch normalization is then applied on the $128\times32$ output \cite{ioffe2015batch}.

In the second layer, a depth-wise $129\times1$ kernel scans over all EEG channels of each temporal filter. This is in contrast to EEGViT's approach, where a kernel of shape $(8,1)$ is used \cite{yang2023vit2eeg}.

Then, similar to EEGViT \cite{yang2023vit2eeg}, the result is passed through a ViT transformer model, with the only difference being the shape of the input data. The base-ViT model \cite{dosovitskiy2020image} was pre-trained on ImageNet-21k and ImageNet 2012 \cite{deng2009imagenet,ridnik2021imagenet} for image classification tasks. EEGViT \cite{yang2023vit2eeg} has previously shown that a ViT model pre-trained for image classification offers surprisingly good results when fine tuned with EEG data.

Lastly, two linear layers on top of the hidden CLS token of the ViT model output the $x,y$ coordinates of predicted gaze position. We have additionally introduced a dropout layer to improve the robustness of the model.

\subsection{Training Parameters and Software Implementation}
We split the EEGEyeNet dataset into 0.7:0.15:0.15 for training, validation, and testing, and the model epoch with the lowest validation RMSE is used for testing. The split is by participant id in the original EEGEyeNet dataset to avoid leakage due to one participant's data samples appearing in more than one of training, validation, testing sets.

We included baseline ML implementations made public by the EEGEyeNet authors to be tested \cite{kastrati2021eegeyenet}. For EEGViT \cite{yang2023vit2eeg}, we ported the authors' implementation match the setup of EEGEyeNet for training and testing in order to have the closest comparisons.

\begin{figure}
    \centering
    \includegraphics[width=0.9\linewidth]{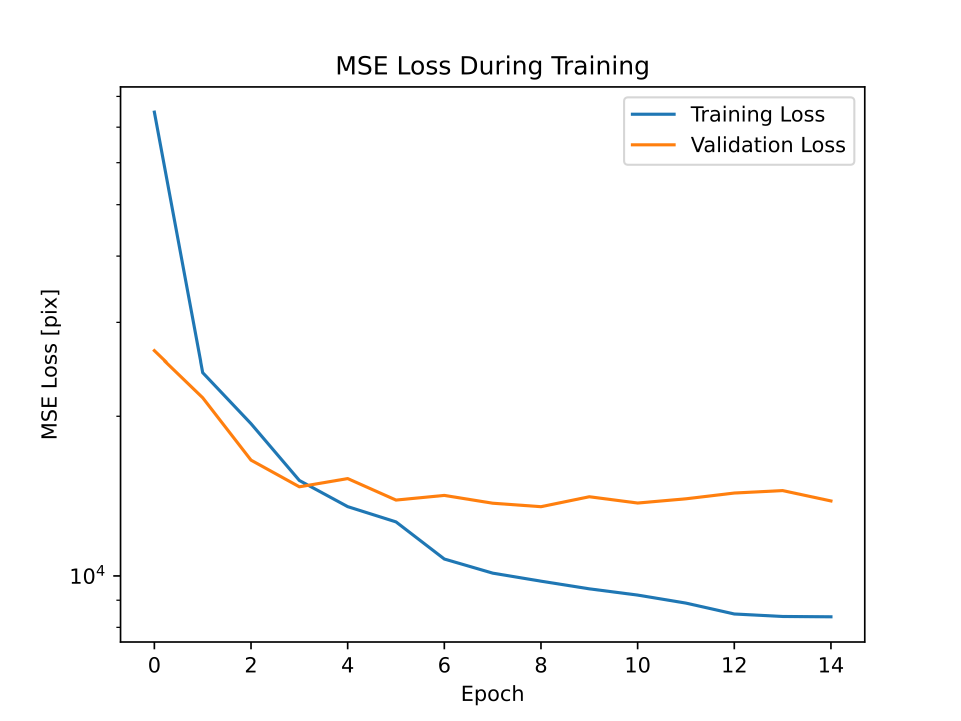}
    \caption{Our Method's MSE Loss During Training}
    \label{fig:loss_during_training}
\end{figure}

\begin{figure}
    \centering
    \includegraphics[width=0.9\linewidth]{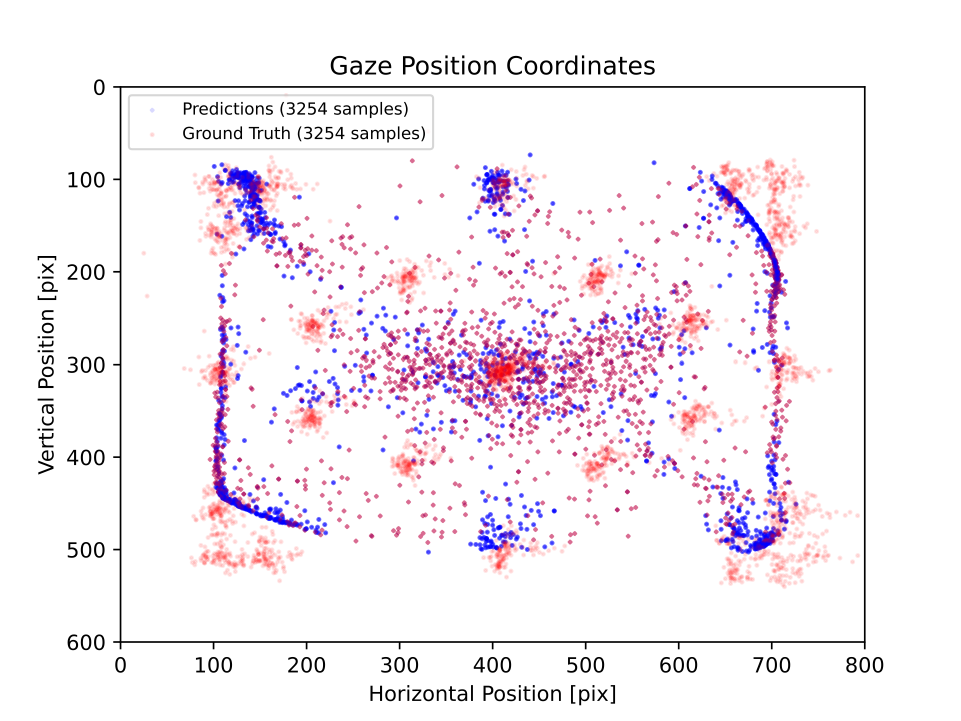}
    \caption{Our method Gaze Position Coordinates, where predictions are colored blue, and the ground truths are colored red \cite{advancing-eeg}.}
    \label{fig:gaze_predictions}
\end{figure}

Our model and EEGViT are trained for 15 epochs in batches of 64 samples, with the Adam Optimizer \cite{kingma2014adam} and an initial learning rate of 1e-4, which is dropped by a factor of 10 every 6 epochs. The model with the lowest validation error is used for testing. An example of the MSE loss during training in one of the runs can be seen in Figure \ref{fig:loss_during_training} and the resulting model's predictions on the testing set can be seen in figure \ref{fig:gaze_predictions}.


The full set of source code can be found at https://github.com/AmCh-Q/CSCI6907Project.

\subsection{Environment Setup}
We performed all training, validation, and testing using Google Colab. We used an Intel Xeon Processor at 2.20GHz, 51 GB of RAM, and a NVIDIA V100 GPU. For CUDA we used version 12.2, for PyTorch we used version 2.1.0, and for Scikit-learn we used version 1.2.2.

\subsection{Evaluation}
EEGEyeNet includes a benchmark where, given samples of shape $(129,500)$ collected from 129 EEG channels at 500Hz for 1 second when the participant fixates on one location, a machine learning model is to be trained to predict the 2-dimensional gaze position (in pixels) of the participant, and the accuracy may be evaluated as either the root mean-squared error (RMSE: Equation \ref{RMSE_eq}) or mean Euclidean distance (MED: Equation \ref{MED_eq}) in pixels or millimeters.
\begin{equation}\label{RMSE_eq}
\text{RMSE}=\sqrt{\frac{\sum_{i=1}^{n}((x_{i, \text{truth}}-x_{i, \text{pred}})^2+(y_{i, \text{truth}}-y_{i, \text{pred}})^2)}{2n}}
\end{equation}
\begin{equation}\label{MED_eq}
\text{MED}=\frac{\sum_{i=1}^{n}\sqrt{(x_{i, \text{truth}}-x_{i, \text{pred}})^2+(y_{i, \text{truth}}-y_{i, \text{pred}})^2}}{n}
\end{equation}
Here $(x_{i, \text{truth}},y_{i, \text{truth}})\in\mathbb{R}^2$ is the coordinates of the gaze position collected with a video-based eye-tracker in the $i$-th sample, and $(x_{i, \text{pred}},y_{i, \text{pred}})\in\mathbb{R}^2$ is the coordinates of the gaze position predicted by machine learning models from EEG data in the $i$-th sample, and $n$ is the number of 1-second data samples collected.

Five runs were run for each of the two metrics above, the mean and standard deviation of the runs were recorded, and the results can be seen in Table \ref{tab:EEGAccuracies} and Table \ref{tab:EEGTimes}.

\begin{table}
    \centering
    \caption{EEGEyeNet Gaze Position Scores and Standard Deviation across 5 Runs of EEGEyeNet baseline methods and EEGViT, compared against our method. Lower is better. All values are in millimeters and rounded to two decimal places. The column "Reported" contains the RMSE values that were originally reported from the respective studies.}
    \label{tab:EEGAccuracies}
    \begin{tabular}{l c c c c}
        \hline
        \textbf{Model} & \textbf{Reported RMSE} & \textbf{Bench RMSE} & \textbf{Bench MED} & \textbf{Study}\\
        \hline
        Naive Center & - & $95.85\pm0$ & $123.43\pm0$ & \cite{kastrati2021eegeyenet}\\
        Naive Mean & $123.3\pm0$ & $95.81\pm0$ & $123.31\pm0$ & \cite{kastrati2021eegeyenet}\\
        Naive Median & - & $95.79\pm0$ & $123.23\pm0$ & \cite{kastrati2021eegeyenet}\\
        \hline
        KNN (K=100) & $119.7\pm0$ & $92.21\pm0$ & $119.67\pm0$ & \cite{kastrati2021eegeyenet}\\
        RBF SVR & $123\pm0$ & $95.56\pm0$ & $123.00\pm0$ & \cite{kastrati2021eegeyenet}\\
        Linear Regression & $118.3\pm0$ & $91.08\pm0$ & $118.37\pm0$ & \cite{kastrati2021eegeyenet}\\
        Ridge Regression & $118.2\pm0$ & $90.91\pm0$ & $118.25\pm0$ & \cite{kastrati2021eegeyenet}\\
        Lasso Regression & $118\pm0$ & $90.80\pm0$ & $118.04\pm0$ & \cite{kastrati2021eegeyenet}\\
        Elastic Net & $118.1\pm0$ & $90.83\pm0$ & $118.13\pm0$ & \cite{kastrati2021eegeyenet}\\
        Random Forest & $116.7\pm0.1$ & $90.09\pm0.08$ & $116.71\pm0.08$ & \cite{kastrati2021eegeyenet}\\
        Gradient Boost & $117\pm0.1$ & $91.01\pm0.06$ & $117.50\pm0.05$ & \cite{kastrati2021eegeyenet}\\
        AdaBoost & $119.4\pm0.1$ & $91.98\pm0.07$ & $119.39\pm0.06$ & \cite{kastrati2021eegeyenet}\\
        XGBoost & $118\pm0$ & $91.73\pm0$ & $118.00\pm0$ & \cite{kastrati2021eegeyenet}\\
        \hline

        CNN & $70.2\pm1.1$ & $59.39\pm0.63$ & $70.11\pm1.56$ & \cite{kastrati2021eegeyenet}\\
        PyramidalCNN & $73.6\pm1.9$ & $60.32\pm1.67$ & $70.86\pm0.87$ & \cite{kastrati2021eegeyenet}\\
        EEGNet & $81.7\pm1.0$ & $61.92\pm0.37$ & $76.93\pm0.73$ & \cite{kastrati2021eegeyenet}\\
        InceptionTime & $70.8\pm0.8$ & $60.32\pm0.74$ & $69.37\pm0.90$ & \cite{kastrati2021eegeyenet}\\
        Xception & $78.7\pm1.6$ & $66.44\pm0.80$ & $76.77\pm1.20$ & \cite{kastrati2021eegeyenet}\\
        \hline
        EEGViT & $55.4\pm0.2$ & $54.41\pm0.76$ & $63.44\pm0.83$ & \cite{yang2023vit2eeg}\\
        \hline
        \textbf{Ours} & - & \textbf{53.06} $\pm 0.73$ & \textbf{60.50} $\pm0.93$ & -\\
        \hline
    \end{tabular}
\end{table}
\begin{table}
    \centering
    \caption{EEGEyeNet Gaze Position run time (model training and validation of 21464 data samples) across 5 runs of EEGEyeNet baseline methods and EEGViT, compared against our method.}
    \label{tab:EEGTimes}
    \begin{tabular}{l c c}
        \hline
        \textbf{Model} & \textbf{Runtime [seconds]} & \textbf{Study}\\
        \hline
        Naive Center & $<0.01$ & \cite{kastrati2021eegeyenet}\\
        Naive Mean & $<0.01$ & \cite{kastrati2021eegeyenet}\\
        Naive Median & $<0.01$ & \cite{kastrati2021eegeyenet}\\
        \hline
        KNN & $0.71 \pm 0.02$ & \cite{kastrati2021eegeyenet}\\
        RBF SVR & $13.23 \pm 0.28$ & \cite{kastrati2021eegeyenet}\\
        LinearReg & $0.40 \pm 0.07$ & \cite{kastrati2021eegeyenet}\\
        Ridge & $0.16 \pm 0.01$ & \cite{kastrati2021eegeyenet}\\
        Lasso & $1.12 \pm 0.02$ & \cite{kastrati2021eegeyenet}\\
        ElasticNet & $1.27 \pm 0.01$ & \cite{kastrati2021eegeyenet}\\
        RandomForest & $355.90 \pm 4.36$ & \cite{kastrati2021eegeyenet}\\
        GradientBoost & $816.69 \pm 6.95$ & \cite{kastrati2021eegeyenet}\\
        AdaBoost & $113.31 \pm 0.08$ & \cite{kastrati2021eegeyenet}\\
        XGBoost & $44.69 \pm 0.43$ & \cite{kastrati2021eegeyenet}\\
        \hline
        CNN & $362.71 \pm 21.52$ & \cite{kastrati2021eegeyenet}\\
        PyramidalCNN & $281.84 \pm 15.27$ & \cite{kastrati2021eegeyenet}\\
        EEGNet & $1696.90 \pm 0.97$ & \cite{kastrati2021eegeyenet}\\
        Xception & $563.30 \pm 10.59$ & \cite{kastrati2021eegeyenet}\\
        \hline
        EEGViT & $2629.97 \pm 5.79$ & \cite{yang2023vit2eeg}\\
        \textbf{Ours} & $812.33 \pm 0.88$ & -\\
        \hline
    \end{tabular}
\end{table}

\section{Discussion}
In this work, we presented an algorithm for predicting gaze position from EEG signals, and Table \ref{tab:EEGAccuracies} shows the comparison of accuracy against various models including the SOTA (EEGViT). As can be seen in both the root mean-squared-error (Equation \ref{RMSE_eq}) and mean euclidean distance (Equation \ref{MED_eq}) metric, our method outperforms the SOTA. This is due to the use of a spatial filtering convolution kernel of shape $(129,1)$, spanning all EEG channels, because the electrode layout, as seen Figure \ref{fig:hydrocel}, appear to be unordered and thus unlikely to be able to be learned through convolution with a smaller kernel as employed by EEGViT, and a kernel spanning all EEG channels would be able to better learn any spatial relationships between any two EEG channels at the same point in time.

We have also inspected the effect of permutation of the EEG channels and found that permuting the order of the EEG channels, either by shuffling or reordering the channels in spiral or z-order, yielded no noticeable difference in accuracy with either our method or EEGViT. We believe this means that the interactions between EEG channel signals is likely too complex and cannot be captured by convolution with a small receptive field.

In addition to measuring the accuracy of the models. We have also measured the run time of each of the models, the result of which are shown in Table \ref{tab:EEGTimes}. While slower than simpler methods such as CNN and even more considerably slower than methods such as KNN or linear regression, our method still offers an approximately 3.2 times speedup compared to the SOTA. This is due to our algorithm utilizing a much large spatial (channel) kernel, reducing the amount of trainable parameters in the model.

We were also able to confirm the findings of EEGEyeNet \cite{kastrati2021eegeyenet} that simple Machine learning models such as KNN, linear regression, and random forest were unable to gather meaningful information from EEG data and yielded no significant difference to naive center (where the model naively predicts the center of the screen), naive mean or naive median (where the model naively predicts the mean or median location of the training set's gaze position), while deep learning models such as CNN and EEGNet were able to yield significantly better results than the naive baselines. We've also discovered that EEGEyeNet may have wrongly reported their results as "root mean-squared-error" when they may have in fact measured the mean euclidean distance error of the models, because in EEGEyeNet's source code we found that they have commented out the codes using RMSE and replaced it with MED, and that the resulting "RMSE" differs significantly with the RMSE result from our experiments, while appearing nearly identical to our "MED" measurements. Since our measured RMSE results on EEGEyeNet's models are significantly lower than reported by the authors of EEGEyeNet, the improvement made from models such as the SOTA, while still noticeable, may be smaller than what may have been believed previously.

\subsection{Limitations}
While the proposed method improves the accuracy and speed compared to the SOTA, the RMSE remains at approximately 5.3 centimeters and the mean euclidean distance remains at 6.1 centimeters, and training and validating the model takes an order of hundreds of seconds. This is considerably worse than commercially available video-based eye-tracking solutions in terms of both accuracy and run time. Moreover, EEGEyeNet's data was recorded in a laboratory setting and the participants were asked to stay still and have their gaze fixated on one spot on a screen, which is not reflective of most real-world application environments of eye-tracking \cite{kastrati2021eegeyenet}. The EEG setup is also more complex than most commercially available video-based solutions. Further research could explore alternative deep learning methods across different datasets for comparative analysis \cite{an2023transfer,an2023survey,jiang2023successfully,gui2024remote,tan2023audio,lu2023machine,ma2022traffic,ma2024data,chen2024trialbench,tan2021multivariate,qiu2023modal,zhao2024deep,zhang2022attention,zhang2023trep}.

\section{Conclusion}
In this paper, we proposed an algorithm of EEG-based gaze prediction that outperforms the SOTA in both accuracy and speed. Our method improves the root mean-squared-error of the tracking to approximately 5.3 centimeters, and we found that having a large depth-wise convolution kernel for all EEG channels had the greatest impact. Nonetheless, EEG-based eye-tracing still has way to go and further research is needed for it to be comparable to the accuracy of traditional video-based eye tracking solutions.

\begin{credits}
\subsubsection{\ackname} This study was part of the authors' work in the course "CSCI 6907 Applied Machine Learning" in The George Washington University.

\subsubsection{\discintname}
The authors declare no competing interests. 
\end{credits}

%
%
%
 \bibliographystyle{splncs04}
 \bibliography{eegeye}

\begin{thebibliography}{10}
\providecommand{\url}[1]{\texttt{#1}}
\providecommand{\urlprefix}{URL }
\providecommand{\doi}[1]{https://doi.org/#1}

\bibitem{altaheri2023deep}
Altaheri, H., Muhammad, G., Alsulaiman, M., Amin, S.U., Altuwaijri, G.A., Abdul, W., Bencherif, M.A., Faisal, M.: Deep learning techniques for classification of electroencephalogram (eeg) motor imagery (mi) signals: A review. Neural Computing and Applications  \textbf{35}(20),  14681--14722 (2023)

\bibitem{an2023transfer}
An, S., Bhat, G., Gumussoy, S., Ogras, U.: Transfer learning for human activity recognition using representational analysis of neural networks. ACM Transactions on Computing for Healthcare  \textbf{4}(1),  1--21 (2023)

\bibitem{an2023survey}
An, S., Tuncel, Y., Basaklar, T., Ogras, U.Y.: A survey of embedded machine learning for smart and sustainable healthcare applications. In: Embedded Machine Learning for Cyber-Physical, IoT, and Edge Computing: Use Cases and Emerging Challenges, pp. 127--150. Springer (2023)

\bibitem{bamatraf2016system}
Bamatraf, S., Hussain, M., Aboalsamh, H., Qazi, E.U.H., Malik, A.S., Amin, H.U., Mathkour, H., Muhammad, G., Imran, H.M.: A system for true and false memory prediction based on 2d and 3d educational contents and eeg brain signals. Computational Intelligence and Neuroscience  \textbf{2016},  45--45 (2016)

\bibitem{chen2024trialbench}
Chen, J., Hu, Y., Wang, Y., Lu, Y., Cao, X., Lin, M., Xu, H., Wu, J., Xiao, C., Sun, J., et~al.: Trialbench: Multi-modal artificial intelligence-ready clinical trial datasets. arXiv preprint arXiv:2407.00631  (2024)

\bibitem{craik2019deep}
Craik, A., He, Y., Contreras-Vidal, J.L.: Deep learning for electroencephalogram (eeg) classification tasks: a review. Journal of neural engineering  \textbf{16}(3),  031001 (2019)

\bibitem{deng2009imagenet}
Deng, J., Dong, W., Socher, R., Li, L.J., Li, K., Fei-Fei, L.: Imagenet: A large-scale hierarchical image database. In: 2009 IEEE conference on computer vision and pattern recognition. pp. 248--255. Ieee (2009)

\bibitem{dosovitskiy2020image}
Dosovitskiy, A., Beyer, L., Kolesnikov, A., Weissenborn, D., Zhai, X., Unterthiner, T., Dehghani, M., Minderer, M., Heigold, G., Gelly, S., et~al.: An image is worth 16x16 words: Transformers for image recognition at scale. arXiv preprint arXiv:2010.11929  (2020)

\bibitem{dou2022time}
Dou, G., Zhou, Z., Qu, X.: Time majority voting, a pc-based eeg classifier for non-expert users. In: International Conference on Human-Computer Interaction. pp. 415--428. Springer (2022)

\bibitem{gao2021complex}
Gao, Z., Dang, W., Wang, X., Hong, X., Hou, L., Ma, K., Perc, M.: Complex networks and deep learning for eeg signal analysis. Cognitive Neurodynamics  \textbf{15}(3),  369--388 (2021)

\bibitem{gui2024remote}
Gui, S., Song, S., Qin, R., Tang, Y.: Remote sensing object detection in the deep learning era—a review. Remote Sensing  \textbf{16}(2), ~327 (2024)

\bibitem{hossain2023status}
Hossain, K.M., Islam, M.A., Hossain, S., Nijholt, A., Ahad, M.A.R.: Status of deep learning for eeg-based brain--computer interface applications. Frontiers in computational neuroscience  \textbf{16},  1006763 (2023)

\bibitem{ioffe2015batch}
Ioffe, S., Szegedy, C.: Batch normalization: Accelerating deep network training by reducing internal covariate shift. In: International conference on machine learning. pp. 448--456. pmlr (2015)

\bibitem{jiang2023successfully}
Jiang, C., Hui, B., Liu, B., Yan, D.: Successfully applying lottery ticket hypothesis to diffusion model. arXiv preprint arXiv:2310.18823  (2023)

\bibitem{kastrati2021eegeyenet}
Kastrati, A., P{\l}omecka, M.B., Pascual, D., Wolf, L., Gillioz, V., Wattenhofer, R., Langer, N.: Eegeyenet: a simultaneous electroencephalography and eye-tracking dataset and benchmark for eye movement prediction. arXiv preprint arXiv:2111.05100  (2021)

\bibitem{key2024advancing}
Key, M.L., Mehtiyev, T., Qu, X.: Advancing eeg-based gaze prediction using depthwise separable convolution and enhanced pre-processing. In: International Conference on Human-Computer Interaction. pp. 3--17. Springer (2024)

\bibitem{advancing-eeg}
Key, M.L., Mehtiyev, T., Qu, X.: Advancing eeg-based gaze prediction using depthwise separable convolution and enhanced pre-processing  (2024), preprint

\bibitem{khan2022transformers}
Khan, S., Naseer, M., Hayat, M., Zamir, S.W., Khan, F.S., Shah, M.: Transformers in vision: A survey. ACM computing surveys (CSUR)  \textbf{54}(10s),  1--41 (2022)

\bibitem{kingma2014adam}
Kingma, D.P., Ba, J.: Adam: A method for stochastic optimization. arXiv preprint arXiv:1412.6980  (2014)

\bibitem{koome2023trends}
Koome~Murungi, N., Pham, M.V., Dai, X., Qu, X.: Trends in machine learning and electroencephalogram (eeg): A review for undergraduate researchers. arXiv e-prints pp. arXiv--2307 (2023)

\bibitem{lawhern2018eegnet}
Lawhern, V.J., Solon, A.J., Waytowich, N.R., Gordon, S.M., Hung, C.P., Lance, B.J.: Eegnet: a compact convolutional neural network for eeg-based brain--computer interfaces. Journal of neural engineering  \textbf{15}(5),  056013 (2018)

\bibitem{li2024enhancing}
Li, W., Zhou, N., Qu, X.: Enhancing eye-tracking performance through multi-task learning transformer. In: International Conference on Human-Computer Interaction. pp. 31--46. Springer (2024)

\bibitem{lu2023machine}
Lu, Y., Shen, M., Wang, H., Wang, X., van Rechem, C., Wei, W.: Machine learning for synthetic data generation: a review. arXiv preprint arXiv:2302.04062  (2023)

\bibitem{ma2022traffic}
Ma, X.: Traffic performance evaluation using statistical and machine learning methods. Ph.D. thesis, The University of Arizona (2022)

\bibitem{ma2024data}
Ma, X., Karimpour, A., Wu, Y.J.: Data-driven transfer learning framework for estimating on-ramp and off-ramp traffic flows. Journal of Intelligent Transportation Systems pp. 1--14 (2024)

\bibitem{murungi2023empowering}
Murungi, N.K., Pham, M.V., Dai, X.C., Qu, X.: Empowering computer science students in electroencephalography (eeg) analysis: A review of machine learning algorithms for eeg datasets. SIGKDD  (2023)

\bibitem{qiu2023modal}
Qiu, Y., Zhao, Z., Yao, H., Chen, D., Wang, Z.: Modal-aware visual prompting for incomplete multi-modal brain tumor segmentation. In: Proceedings of the 31st ACM International Conference on Multimedia. pp. 3228--3239 (2023)

\bibitem{qu2022time}
Qu, X.: Time Continuity Voting for Electroencephalography (EEG) Classification. Ph.D. thesis, Brandeis University (2022)

\bibitem{qu2019personalized}
Qu, X., Hall, M., Sun, Y., Sekuler, R., Hickey, T.J.: A personalized reading coach using wearable eeg sensors. CSEDU  (2019)

\bibitem{qu2022eeg4home}
Qu, X., Hickey, T.J.: Eeg4home: A human-in-the-loop machine learning model for eeg-based bci. In: Augmented Cognition: 16th International Conference, AC 2022, Held as Part of the 24th HCI International Conference, HCII 2022, Virtual Event, June 26--July 1, 2022, Proceedings. pp. 162--172. Springer (2022)

\bibitem{qu2020multi}
Qu, X., Liu, P., Li, Z., Hickey, T.: Multi-class time continuity voting for eeg classification. In: Brain Function Assessment in Learning: Second International Conference, BFAL 2020, Heraklion, Crete, Greece, October 9--11, 2020, Proceedings 2. pp. 24--33. Springer (2020)

\bibitem{qu2020identifying}
Qu, X., Liukasemsarn, S., Tu, J., Higgins, A., Hickey, T.J., Hall, M.H.: Identifying clinically and functionally distinct groups among healthy controls and first episode psychosis patients by clustering on eeg patterns. Frontiers in psychiatry  \textbf{11},  541659 (2020)

\bibitem{qu2020using}
Qu, X., Mei, Q., Liu, P., Hickey, T.: Using eeg to distinguish between writing and typing for the same cognitive task. In: Brain Function Assessment in Learning: Second International Conference, BFAL 2020, Heraklion, Crete, Greece, October 9--11, 2020, Proceedings 2. pp. 66--74. Springer (2020)

\bibitem{qu2018eeg}
Qu, X., Sun, Y., Sekuler, R., Hickey, T.: Eeg markers of stem learning. In: 2018 IEEE Frontiers in Education Conference (FIE). pp.~1--9. IEEE (2018)

\bibitem{ridnik2021imagenet}
Ridnik, T., Ben-Baruch, E., Noy, A., Zelnik-Manor, L.: Imagenet-21k pretraining for the masses. arXiv preprint arXiv:2104.10972  (2021)

\bibitem{roy2019deep}
Roy, Y., Banville, H., Albuquerque, I., Gramfort, A., Falk, T.H., Faubert, J.: Deep learning-based electroencephalography analysis: a systematic review. Journal of neural engineering  \textbf{16}(5),  051001 (2019)

\bibitem{schirrmeister2017deep}
Schirrmeister, R.T., Springenberg, J.T., Fiederer, L.D.J., Glasstetter, M., Eggensperger, K., Tangermann, M., Hutter, F., Burgard, W., Ball, T.: Deep learning with convolutional neural networks for eeg decoding and visualization. Human brain mapping  \textbf{38}(11),  5391--5420 (2017)

\bibitem{tan2021multivariate}
Tan, J., Shen, X., Zhang, X., Wang, Y.: Multivariate encoding analysis of medial prefrontal cortex cortical activity during task learning. In: 2021 43rd Annual International Conference of the IEEE Engineering in Medicine \& Biology Society (EMBC). pp. 6699--6702. IEEE (2021)

\bibitem{tan2023audio}
Tan, J., Zhang, X., Wu, S., Song, Z., Chen, S., Huang, Y., Wang, Y.: Audio-induced medial prefrontal cortical dynamics enhances coadaptive learning in brain--machine interfaces. Journal of Neural Engineering  \textbf{20}(5),  056035 (2023)

\bibitem{vaswani2017attention}
Vaswani, A., Shazeer, N., Parmar, N., Uszkoreit, J., Jones, L., Gomez, A.N., Kaiser, {\L}., Polosukhin, I.: Attention is all you need. Advances in neural information processing systems  \textbf{30} (2017)

\bibitem{wang2022eeg}
Wang, R., Qu, X.: Eeg daydreaming, a machine learning approach to detect daydreaming activities. In: International Conference on Human-Computer Interaction. pp. 202--212. Springer (2022)

\bibitem{wolf2022deep}
Wolf, L., Kastrati, A., P{\l}omecka, M.B., Li, J.M., Klebe, D., Veicht, A., Wattenhofer, R., Langer, N.: A deep learning approach for the segmentation of electroencephalography data in eye tracking applications. arXiv preprint arXiv:2206.08672  (2022)

\bibitem{xiang2022too}
Xiang, B., Abdelmonsef, A.: Too fine or too coarse? the goldilocks composition of data complexity for robust left-right eye-tracking classifiers. arXiv preprint arXiv:2209.03761  (2022)

\bibitem{xiang2022vector}
Xiang, B., Abdelmonsef, A.: Vector-based data improves left-right eye-tracking classifier performance after a covariate distributional shift. In: International Conference on Human-Computer Interaction. pp. 617--632. Springer (2022)

\bibitem{yang2023vit2eeg}
Yang, R., Modesitt, E.: Vit2eeg: Leveraging hybrid pretrained vision transformers for eeg data. arXiv preprint arXiv:2308.00454  (2023)

\bibitem{yi2022attention}
Yi, L., Qu, X.: Attention-based cnn capturing eeg recording’s average voltage and local change. In: Artificial Intelligence in HCI: 3rd International Conference, AI-HCI 2022, Held as Part of the 24th HCI International Conference, HCII 2022, Virtual Event, June 26--July 1, 2022, Proceedings. pp. 448--459. Springer (2022)

\bibitem{zhang2023trep}
Zhang, Z., Tian, R., Ding, Z.: Trep: Transformer-based evidential prediction for pedestrian intention with uncertainty. In: Proceedings of the AAAI Conference on Artificial Intelligence. vol.~37, pp. 3534--3542 (2023)

\bibitem{zhang2022attention}
Zhang, Z., Tian, R., Sherony, R., Domeyer, J., Ding, Z.: Attention-based interrelation modeling for explainable automated driving. IEEE Transactions on Intelligent Vehicles  \textbf{8}(2),  1564--1573 (2022)

\bibitem{zhao2024deep}
Zhao, S., Yang, X., Zeng, Z., Qian, P., Zhao, Z., Dai, L., Prabhu, N., Nordlund, P., Tam, W.L.: Deep learning based cetsa feature prediction cross multiple cell lines with latent space representation. Scientific Reports  \textbf{14}(1), ~1878 (2024)

\bibitem{zhou2022brainactivity1}
Zhou, Z., Dou, G., Qu, X.: Brainactivity1: A framework of eeg data collection and machine learning analysis for college students. In: International Conference on Human-Computer Interaction. pp. 119--127. Springer (2022)

\end{thebibliography}

\end{document}